# Parameter Compression of Recurrent Neural Networks and Degradation of Short-term Memory

Jonathan A. Cox
Qualcomm Technologies, Inc.
San Diego, United States
joncox@alum.mit.edu

*Abstract*—The significant computational costs of deploying neural networks in large-scale or resource constrained environments, such as data centers and mobile devices, has spurred interest in model compression, which can achieve a reduction in both arithmetic operations and storage memory. Several techniques have been proposed for reducing or compressing the parameters for feed-forward and convolutional neural networks, but less is understood about the effect of parameter compression on recurrent neural networks (RNN). In particular, the extent to which the recurrent parameters can be compressed and the impact on short-term memory performance, is not well understood. In this paper, we study the effect of complexity reduction, through singular value decomposition rank reduction, on RNN and minimal gated recurrent unit (MGRU) networks for several tasks. We show that considerable rank reduction is possible when compressing recurrent weights, even without fine tuning. Furthermore, we propose a perturbation model for the effect of general perturbations, such as a compression, on the recurrent parameters of RNNs. The model is tested against a noiseless memorization experiment that elucidates the short-term memory performance. In this way, we demonstrate that the effect of compression of recurrent parameters is dependent on the degree of temporal coherence present in the data and task. This work can guide on-the-fly RNN compression for novel environments or tasks, and provides insight for applying RNN compression in low-power devices, such as hearing aids.

*Keywords—compression; recurrent neural networks; complexity reduction; SVD; RNN; MGRU; MRU*

## I. INTRODUCTION

There has been considerable interest in the deployment of artificial neural network models for a wide range of applications, including biomedical devices [1], drones, mobile phones [2] and autonomous vehicles [3]. However, such models are often extremely computationally complex, and can have hundreds of millions of parameters for both recurrent and convolutional neural networks [3], [4]. Reducing the complexity of neural networks is imperative, not only for improved efficiency, but for enabling novel applications in resource constrained environments, such as for hearing aids. Recent progress in complexity reduction, or compression, of feed-forward neural networks has demonstrated reductions of at least 5-10x for the parameters of fully connected layers. Related methods applied to convolutional neural networks (CNN) have demonstrated a reduction of 3-5x [5]–[7].

Recurrent neural networks are especially interesting for low-power and mobile applications since they can often involve real-time processing of sequential information. Although there has been work investigating pruning, compression and rank reduction of feed-forward and convolutional neural networks, it is not well understood how complexity reduction impacts recurrent neural networks (RNN), and memory-cell based architectures such as the long short-term memory (LSTM), gated recurrent unit (GRU) or the recently proposed minimal gated recurrent unit (MGRU) [8]–[10]. While work by Geras et al. [11] has investigated using model compression to train a CNN from an LSTM, it is not always possible or desirable to transform an RNN into a CNN for practical and theoretical reasons (for instance, when very long-range time dependencies are inherent in the task). Nevertheless, there is significant interest in complexity reduction of RNNs, as they have witnessed large-scale adoption in industrial systems [12], [13].

In contrast to feed-forward neural networks, such as CNNs, information can flow through the recurrent feedback connections of an RNN an indeterminate number of cycles. In general, during inference, it is often not known in advance how long an RNN must be unfolded, such as during image captioning or object detection [14], [15]. While this capability makes RNNs extremely powerful and expressive, applying and understanding complexity reduction is more challenging [16]. As a result, compression of RNN parameters becomes dependent on the temporal dependencies embedded in the data and task, which may not be fully known during inference and can change over time.

In this paper, we show that recurrent neural networks, including those using a memory cell based architecture, such as MGRU, achieve significant complexity reduction of the feed-forward and recurrent connection weights, for both classification and language modeling sequence prediction tasks. In addition, we provide a more fundamental understanding of how complexity reduction, viewed as a general perturbation or corruption, is impacted by temporal dependency. Therefore, we devise a perturbation model of the effect of a general compression method, such as singular value decomposition (SVD) rank reduction, on the short-term memory performance of recurrent networks. This model is tested on a noiseless memorization task to elucidate the conditions over which scaling of short-term memory performance agrees. In this way, it is shown how the achievable compression is dependent on the degree of temporal coherence present in the task and data.

## II. RANK REDUCTION OF RECURRENT NEURAL NETWORKS

Long short-term memory (LSTM) networks have been extremely popular due to their considerable practical success. However, it has recently been shown that there is redundancy in the LSTM structure, which has led to new architectures, such as the GRU [10]. There has also been renewed interest in RNNs owing to more powerful optimization algorithms, such as Hessian-free optimization [17]. It has been suggested that the core attribute of the LSTM is the memory cell architecture, and that comparable performance is obtained with fewer [18], and possibly even a single, flow control gate, as for the MGRU [8]. Therefore, to best capture the underlying dynamics, our analysis is performed with RNN and MGRU architectures. The MGRU represents the most fundamental incarnation of a gated differentiable memory cell (GDMC), and is valuable for understanding how compression impacts both RNN and GDMC-based networks.

A standard recurrent unit (RNN) is an artificial neural network with recurrent, or feedback, connections within a fully connected layer, as in Eqn. (1).

$$a(t) = f\left[W \cdot a^{(l-1)}(t) + W_r \cdot a(t-1)\right] \quad (1)$$

Here, $f$ is a general non-linear function, such as the hyperbolic tangent or rectifying nonlinearity (ReLU), $W$ is the matrix of feed-forward connections, $W_r$ is the matrix of recurrent connections and $a^{(l-1)}$ are the activations from the previous layer. The recurrent matrix transforms the output of the RNN layer from the previous time step, $t$.

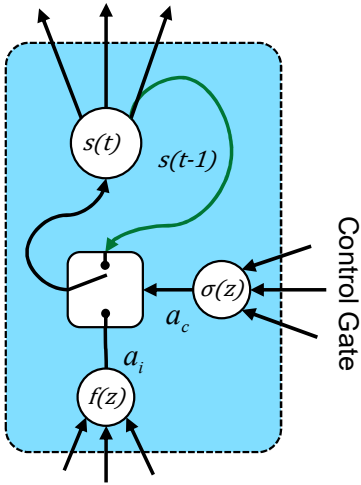

Fig. 1 The minimal gated recurrent unit (MGRU) cell is the simplest gated differentiable memory cell architecture (GDMC), consisting of a single control gate for a switch that controls the storage of information in the cell. In contrast to the LSTM, the switch prevents the memory cell value from exploding, and reduces the need for an additional output nonlinearity. The recurrent connection for the switch is shown by a green arrow.

The minimal gated recurrent unit (MGRU) is a recently proposed reduction of the GRU, as shown in Fig. 1. The MGRU has only a single gate for controlling the memory cell state. In contrast to the basic RNN, differentiable memory cell architectures demonstrate superior long short-term memory performance on a variety of tasks. Consequently, it is reasonable to presume that they may be less susceptible to perturbations in the recurrent weights, such as from rank reduction, since information could remain in the cell state with less influence due to successive perturbations from the recurrent weights. The MGRU, as described by Eqn. (2), differs from the RNN in that there is an additional gate that "switches" the output state between a linear combination of the input and the prior state, s(t-1). In this way, it represents the simplest form of differentiable memory cell, and lacks several features of the LSTM that add complexity, such as: an output nonlinearity, an input gate, an output gate and peephole connections. In Eqn. (2), σ represents the sigmoid function, while $W_i^f$ and $W_c^f$ are the forward connections for the input and control units, respectively. Similarly, $W_i^r$ and $W_c^r$ represent the recurrent connections. In practice, we combine the forward and recurrent matrices into single matrices for forward and recurrent connections.

$$a_c(t) = \sigma\left[W_c^f \cdot a^{(l-1)}(t) + W_c^r \cdot s(t-1)\right]$$
$$a_i(t) = f\left[W_i^f \cdot a^{(l-1)}(t) + W_i^r \cdot s(t-1)\right] \quad (2)$$
$$s(t) = a_c(t) \cdot a_i(t) + (1 - a_c(t))s(t-1)$$

An effective form of complexity reduction, which has been demonstrated on feed-forward and convolutional neural networks, is rank reduction via singular value decomposition on the network parameters. For RNNs, the forward and recurrent matrix of weights can be individually decomposed into their singular values and orthonormal bases, $\Sigma$ and U, V, respectively. By eliminating the smallest singular values, in order from least to greatest, an optimal reduced rank representation, $\tilde{Q}$ and $\tilde{V}$ is found, as in Eqn. (3). This compressed representation has only R·(M+N) parameters, where R is the rank and M and N are the original dimensions of a particular weight matrix. The rank reduced matrix-vector product can be viewed as decomposing a single layer into two linear layers, parameterized by $\tilde{Q}$ and $\tilde{V}$, such that the computation is performed as in (4), where $x_0$ is the bias and $x$ is the vector of gate inputs.

$$W = U\Sigma V^T \approx (\tilde{U}\tilde{\Sigma})\tilde{V}^T \triangleq \tilde{Q}\tilde{V}^T \quad (3)$$

$$a = f(Wx + x_0) \approx f(\tilde{Q}(\tilde{V}^T x) + x_0) \quad (4)$$

## III. PERTUBATION MODEL FOR SHORT-TERM MEMORY

One of the strengths of recurrent neural networks is the ability to learn temporal sequence tasks requiring some degree of short-term memory capability that is learned directly from the data. In contrast to feed-forward networks, the recurrent neural network is also unfolded into a deep network in time, with shared recurrent weights at every time step. In this way, information is repeatedly transformed by the recurrent weights. As a result, they are especially sensitive to corruptions and perturbations, which is why performance suffers when Dropout is applied naively to the recurrent connections [19]–[21]. For long sequence problems, information is corrupted by the perturbation to the recurrent weights over many time steps. Consequently, it is reasonable to suspect that recurrent connections cannot benefit significantly from compression. Compounding the problem, the propagation length of information in an unfolded RNN is unknown at inference time and dependent on the specific task and data encountered. We therefore wish to describe a perturbation model for understanding how general perturbations to the recurrent weights, such as SVD rank reduction, impact the fundamental performance scaling of recurrent networks.

Consider a standard RNN with tanh activation nonlinearity, which is biased near zero and has small activations, perhaps through a sparsity activation penalty [22]. In this case, it is reasonable to linearize the activation function (for the purposes of our analysis), within some regime, as in Eqn. (5). Furthermore, we can simplify the effect of an arbitrary compression scheme, such as SVD rank reduction, as a perturbation $\delta$ on the original weight matrix, $\tilde{W} = \tilde{Q}\tilde{V}^T = W + \delta$, where $\delta \ll 1$.

$$f(z) = \tanh(z) \approx z \approx (W_r + \delta)x \quad (5)$$

To understand the effect of a small perturbation on short-term memory performance, consider the noiseless memorization experiment described by Martens and Sutskever [17] and shown in Fig. 2. For this task, an RNN is presented with a sequence of bits, $N_b$ long, while it is unfolded over $T$ time-steps. At $t = T-N_b$, the network is asked to reproduce the bit sequence that was initially presented when the stop-bit, $s$, is presented. This task elucidates the short-term memory performance scaling of a network and allows us to model the degradation due to a perturbation, such as complexity reduction, and gain fundamental insight.

Performance on this task is evaluated by the difference in the ground-truth output and the actual output, $\Delta b$, after $T$ successive unfoldings. Since the input and output weights, $W_i$ and $W_o$, are unperturbed we neglect them for simplicity. After neglecting higher order terms of $\delta$ in Eqn. (6), we have a model for the error due to the effect of a perturbation. Clearly, in the regime where the assumptions remain valid, the error scales linearly with the temporal coherence, $T$, and the magnitude of the perturbation $\delta$. Also, the spectral radius of the recurrent weight matrix, $\rho$, should be set $\rho(W_r) \leq 1$ so that the error does not blow up. Nevertheless, there is a tradeoff between the desire to set $\rho(W_r) > 1$ to encourage short-term memory and to reduce the amplification of error due to a perturbation [23].

$$\Delta b = b - \tilde{b}$$
$$= (W_r + \delta)^T x - (W_r)^T x \quad (6)$$
$$\approx T\delta(W_r)^{T-1} x$$

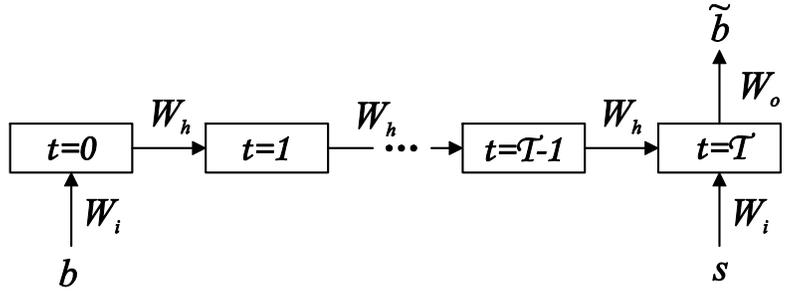

Fig. 2. The unfolded RNN for the noiseless memorization experiment as described in [17]. Initially, a sequence of bits, b, are presented to the network, at which point it is unfolded over T time-steps. After unfolding, a stop-word s is presented, which asks the network to reproduce the bits that were initially present. Performance on the task elucidates the scaling of short-term memory performance.

Based on this model, we expect that RNN complexity reduction on network performance is dependent on the degree of temporal coherence in the data and task, which is often determined by the environment and stochastic. In general, it is not known at inference and is not stationary. The examples of image caption generation and object detection in crowded scenes illustrate this point [14], [15], since the output sequence is highly dependent on the input data found in the field.

## IV. EXPERIMENTS

Three separate experiments are performed on three different tasks for both RNN and MGRU networks. The goal of the first two experiments is to understand how complexity reduction through SVD rank reduction separately impacts performance for feed-forward and recurrent connections. In particular, we are interested in real-time complexity reduction of RNNs without the benefit of additional fine-tuning. Lastly, we perform the noiseless memorization experiment to understand the regime over which the perturbation model is applicable, and how it applies to RNN and MGRU networks. In all three models with MGRU cells, SVD compression is applied to a single matrix of all recurrent connections, including those to the control gates.

### A. Language Model

In the first experiment, we train a recurrent language model to predict the next word in a sequence by minimizing the cross-entropy error over the full vocabulary, as described in [19]. We use the complete works of Shakespeare as a corpus for this task, and apply the Stanford Treebank Tokenizer (PTBTokenizer) library to tokenize the corpus [24]. The resulting corpus has a vocabulary of 26,430 words. All models and experiments consist of a single recurrent layer and are trained using the Adam optimizer [25] with Dropout applied to the hidden layer outputs. For the language modeling task, we use a batch size of 20 (by dividing the corpus into equal portions) and train the network via continuous, ordered passes through the corpus for 30 epochs. Both the RNN and MGRU networks have a single recurrent layer with 500 units, which is fed by a word embedding matrix of 500 dimensions per word. The performance on this task is measured as the mean perplexity over the full vocabulary distribution.

After training, we construct a new, lower rank model with SVD compressed parameters. Both the feed-forward weights that are incoming to the recurrent layer and the recurrent connections within the recurrent layer are compressed. The ranks are separately swept from 1 to 500 (full rank) as the perplexity is recorded over an entire epoch (see Fig. 3). No fine-tuning is performed after rank reduction.

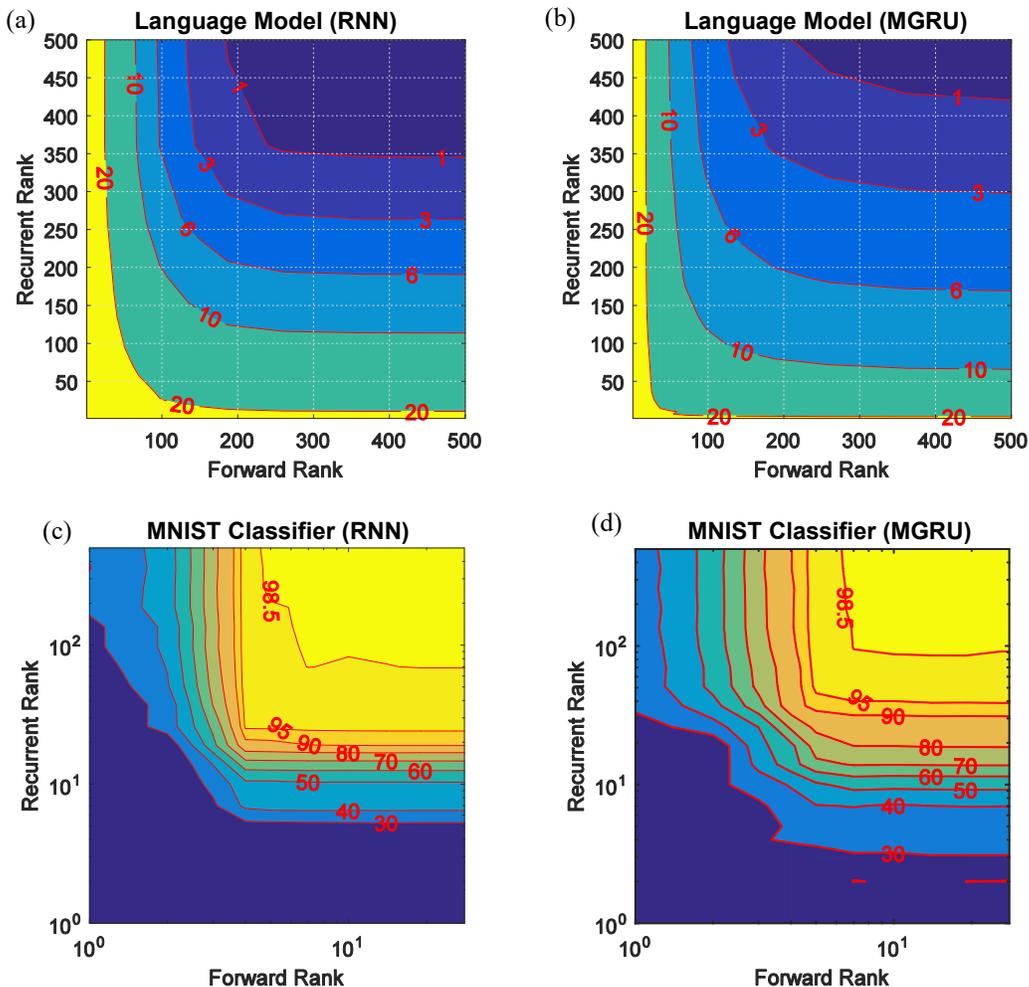

Fig. 3. The above plots show the performance on a task (perplexity or accuracy) verses the rank of the recurrent and forward matricies, after SVD rank reduction, by eliminating the smallest singular values in order from least to greatest. For RNN models, "recurrent rank" is the rank of the Wr matrix, and "forward rank" is the rank of the W matrix in Eqn (1). For MGRU models, "recurrent rank" is the rank of a combined $W_t^r$ and $W_c^r$ matrix. Similary, "forward rank" is the rank of a combined $W_t^f$ and $W_c^f$ matrix. (a)-(d) show the logarithmically scaled performance verses the rank of the feed-forward and recurrent connections RNN or MGRU models. (a)-(b) are the log perplexity on a language modeling task with isolines shown at dB levels (1 dB ≈ 12%). (c)-(d) show the performance on MNIST classification with units of accuracy (%), and with log scaling for rank.

The isolines of the contour plots in Fig. 3 are shown for a logarithmic increase in perplexity, $20\log_{10}(P/P_{\min})$. Thus, a 1 dB increase in perplexity is approximately 12%. For this experiment, relatively greater rank reduction is possible for feed-forward connections than for recurrent connections, by a factor of about 2:1—for both the RNN and MGRU models. Without fine tuning, significant rank reduction is possible with minor degradation in performance. Moreover, we have observed that practical models (see [19]) have even greater redundancy, and tend to be highly over-parameterized in comparison to this simplified example.

*B. MNIST Classifier*

The second experiment is performed with a single-layer recurrent MNIST classifier. In this case, the data is presented to the RNN as one 28-dimension column vector per time-step, over 28 time-steps. In effect, the RNN observes the image one "scan line" at a time and must make sense of the total image. The output of the recurrent hidden layer, which has also 500 units, is temporally mean-pooled and sent to a fully-connected output layer with softmax activation and cross-entropy loss over the 10 classes. The rank of the feed-forward connections is at most 28, which is the dimensionality of the input, while it is at most 500 for the recurrent connections. In contrast to the language model, the rank of both feed-forward and recurrent weights is reduced by similar ratios of about 6x, along the 98.5% isoline. However, as we shall see from the next experiment, the degree of reduction is dependent on the task and data, and it may not always be possible to achieve significant compression in the recurrent weights when long short-term memory performance is critical. Understanding this tradeoff is imperative when deploying real-world models in the field.

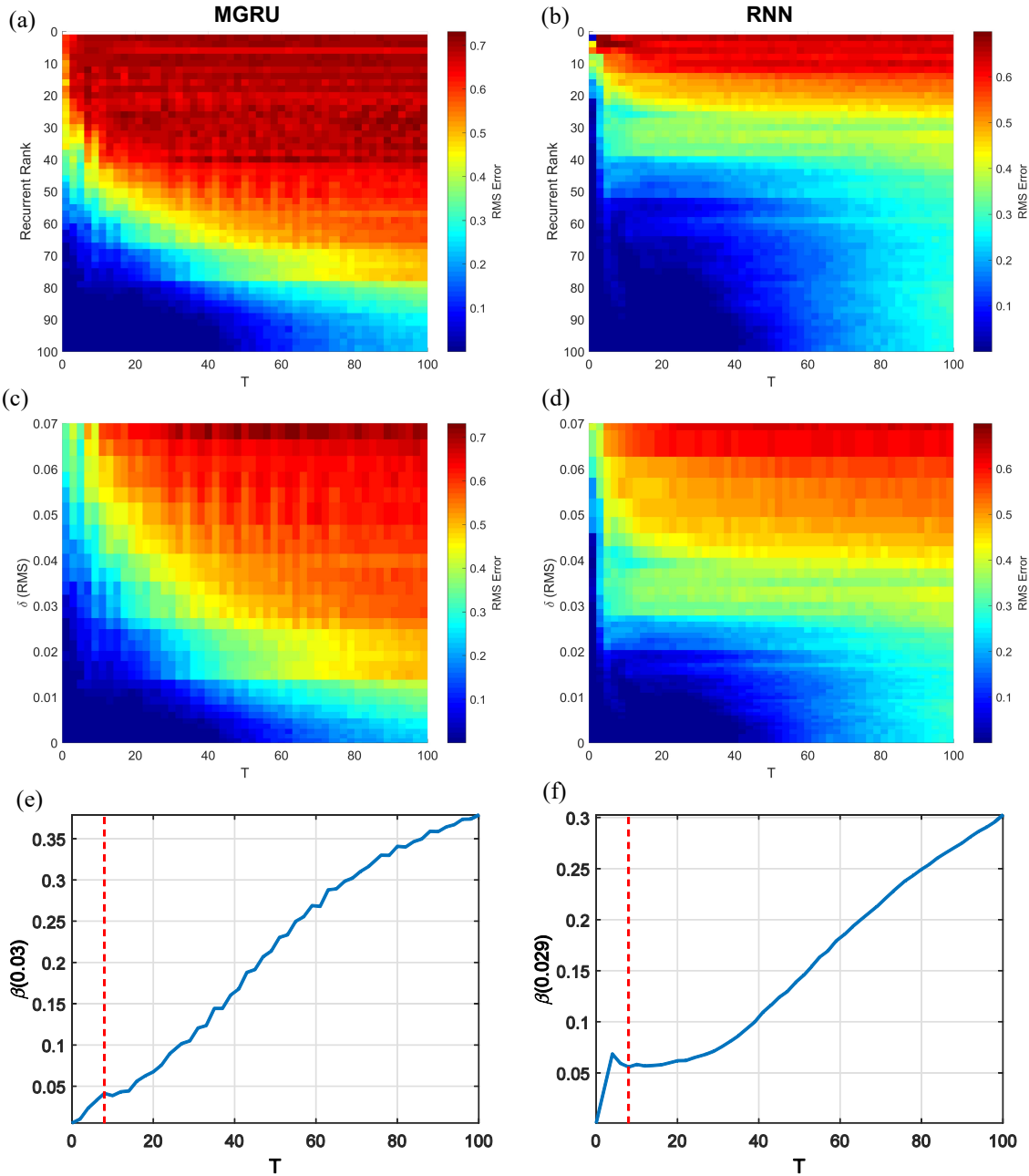

Fig. 4. The performance in root mean square (RMS) error for the noiseless memorization task for RNN (left column) and MGRU (right column) networks. (a)-(b) show the error over recurrent connect rank vs. temporal delay, T. In (c)-(d), the vertical axis is rescaled from rank to equivalent RMS error, $\delta$, for the approximated recurrent weight matrix. Note, $\delta \approx 0.07$ corresponds to ranks of 39 and 11 for RNN and MGRU, respectively. Lastly, (e)-(f) show a collapsed view of (c)-(d), where the error is integrated over the vertical axis from 0 to $\delta \approx 0.03$, (ranks of 64 and 37, respectively).

### C. Noiseless Memorization and Temporal Coherence

The final experiment was conducted to verify the performance scaling of the perturbation model for short-term memory. To this end, we trained RNN and MGRU noiseless memorization models as in Fig. 2, and described in [17], with 100 recurrent hidden units. In both cases, a sequence of $N_b = 8$ bits drawn from a Bernoulli distribution with $p = 0.5$ are presented to the network. After a "silent" period of 0 to 30 steps is observed, a stop-word is presented and the network must recall the sequence. A period of $T = 0$ corresponds to the trivial case where the network must immediately reproduce the input within a single time step (no memorization). Similarly, $T = 1$ would ask the network to remember for only a single time step. Both RNN and MGRU networks are trained with mini-batches of 64 where $T \in [0, 30]$ is randomly drawn, once per batch. Both RNN and MGRU models fully converge, however MGRU does so much faster.

After training, the rank of the recurrent connections is swept for various values of $T$, as shown in Fig. 4. Each point on the surface is the average of 1000 trials. For large $\delta$, the approximation breaks down, and the error saturates almost immediately. Here, $\delta$ is the root mean squared (RMS) error from a rank R approximation of the recurrent weight matrix. To estimate how the error, $\Delta b$, scales with duration, $T$, the 2D plot is collapsed by integrating over $\delta$ in the linear regime (for small perturbation). Thus, $\beta$ in Eqn. (7) is the mean integrated RMS error up to some peak perturbation, at which point the perturbation model is no longer in the valid regime.

$$\beta(\delta_f) = \frac{1}{N_\delta} \sum_{d=0}^{\delta_f} \sqrt{\frac{1}{T} \sum_{t=0}^{T} \Delta b(d,t)^2} \qquad (7)$$

The results shown on the bottom row of Fig. 4 confirm that the error scales linearly with $\delta$ and $T$, supporting the proposed perturbation model. Interestingly, the memory cell architecture, or MGRU, exhibits similar behavior, but is less sensitive when $T$ is small. This may indicate that for moderate temporal durations (T<30) the MGRU is able to accurately retain short-term memory in the cell state without subjecting it to repeated perturbations. However, beyond this duration, it becomes more sensitive to degradation, perhaps because the model was trained with durations of T up to 30. In contrast, RNN error scales quite linearly, even for short durations of $T$.

## V. Conclusion

Our results demonstrate that both the feed-forward and recurrent connections of RNN and differentiable memory-cell architectures (MGRU) benefit from parameter compression, which has considerable practical benefit for low-power and resource constrained operating environments. Unlike for strictly feed-forward networks, such as CNNs, compression of recurrent connections impacts performance in the temporal domain, which is dependent on the sequential coherence in the data and task. This temporal dependency is often unknown until during inference, and may vary over time. Results suggest that MGRU is less sensitive to recurrent parameter compression when faced with varying temporal depenence in the data. Finally, we proposed and experimentally validated a pertubation model governing the scaling of short-term memory performance due to parameter compression. Consequently, this work will guide real-time RNN compression for practical applications, when deploying trained models in the field. For instance, by estimating the temporal coherence of the data and adjusting the compression in real-time, minimal resource utilization can be achieved for applications ranging from hearing aids to mobile devices.